\documentclass[journal]{IEEEtran}

\ifCLASSINFOpdf
\else
   \usepackage[dvips]{graphicx}
\fi
\usepackage{url}

\usepackage{graphicx}
\usepackage{stfloats}

\begin{document}

\title{Automatically Extract the Semi-transparent Motion-blurred Hand from a Single Image}

\author{ Xiaomei Zhao, Yihong Wu
\thanks{This work was supported by the National Natural Science Foundation of China under 61836015 , 61421004 , 61572499.}
\thanks{X. Zhao is with the National Laboratory of Pattern Recognition, Institute of Automation, Chinese Academy of Sciences, Beijing 100190, China, and also with the University of Chinese Academy of Sciences, Beijing, China.}
\thanks{Y. Wu is with the National Laboratory of Pattern Recognition, Institute of Automation, Chinese Academy of Sciences, Beijing 100190, China, and also with the University of Chinese Academy of Sciences, Beijing, China (e-mail: yhwu@nlpr.ia.ac.cn).}}

\maketitle

\begin{abstract}

When we use video chat, video game, or other video applications, motion-blurred hands often appear. Accurately extracting these hands is very useful for video editing and behavior analysis. However, existing motion-blurred object extraction methods either need user interactions, such as user supplied trimaps and scribbles, or need additional information, such as background images. In this paper, a novel method which can automatically extract the semi-transparent motion-blurred hand just according to the original RGB image is proposed. The proposed method separates the extraction task into two subtasks: alpha matte prediction and foreground prediction. These two subtasks are implemented by Xception based encoder-decoder networks. The extracted motion-blurred hand images can be calculated by multiplying the predicted alpha mattes and foreground images. Experiments on synthetic and real datasets show that the proposed method has promising performance.

\end{abstract}


\IEEEpeerreviewmaketitle

\section{Introduction}

\IEEEPARstart{H}{and} language is one of the most important human gesture languages. Poor hand extraction results can greatly reduce the performance of image or video editing, behavior recognition, and behavior analysis. Extracting hands can be implemented by hand segmentation methods [1, 2]. However these methods can’t deal with motion-blurred hands, which are very common in practical applications. Existing traditional methods [3-5], which were designed to predict the alpha mattes or foreground images of motion-blurred objects, generally need user interactions [3, 4] or short-exposure frames [5]. Zhao et. al. [6] proposed a deep learning network to predict the alpha mattes of motion-blurred hands, and then extracted motion-blurred hands by subtracting background components from the original images. A simple flow chart of this method is shown in Fig. 1 (a). An obvious drawback of this method is that it needs background images, which are inconvenient to obtain. In this paper, we propose a method which can automatically extract semi-transparent motion-blurred hands just according to the original RGB images, without requiring any additional information.

\begin{figure}
\centerline{\includegraphics[width=\columnwidth]{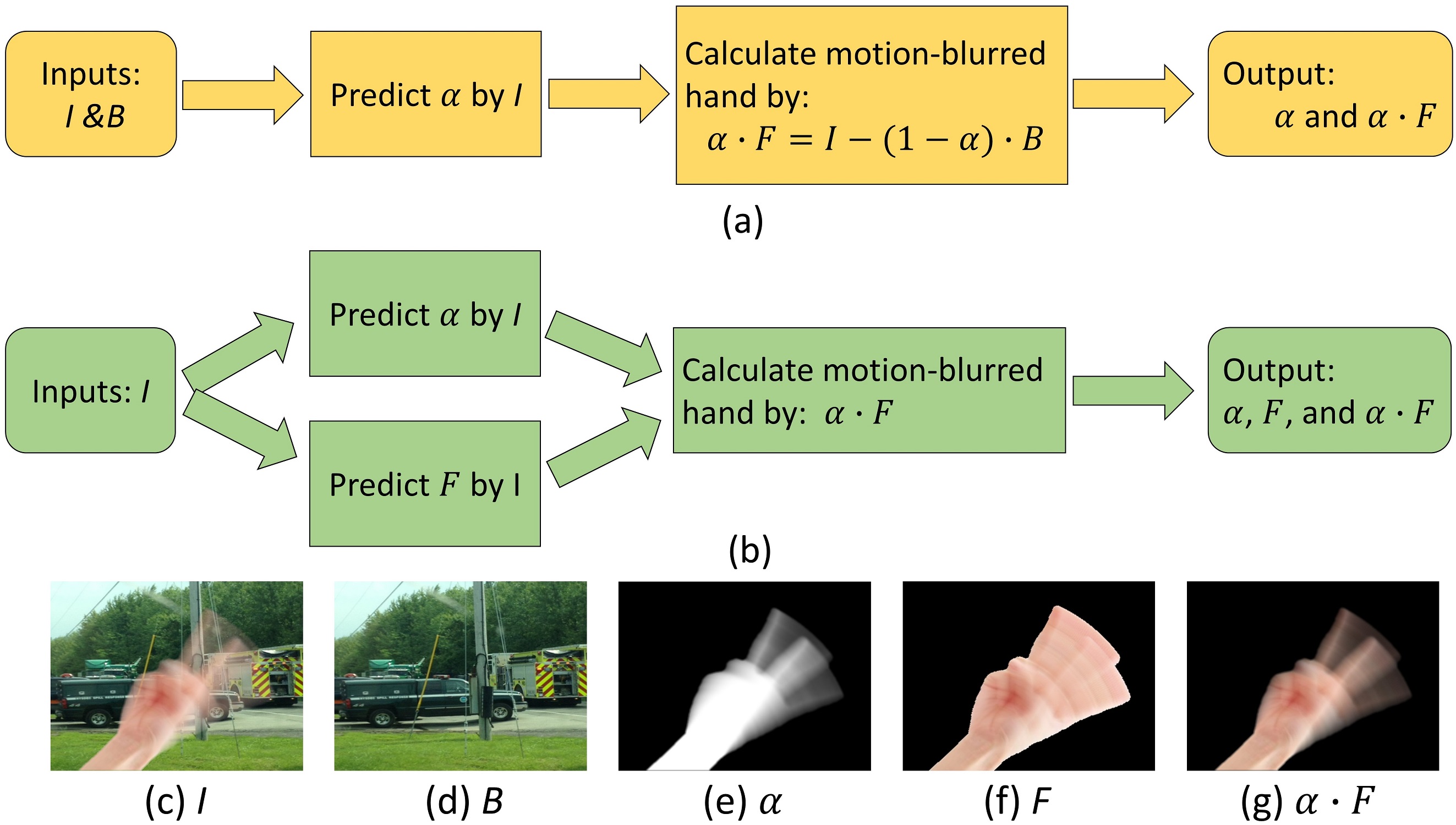}}
\caption{ Compare the flow charts of different motion-blurred hand extraction methods: (a) the method proposed in [6]; (b) the method proposed in this paper. (c-g) show examples of different kinds of images. In this figure, $I$ denotes original image, $B$ denotes background image, $F$ denotes foreground image, $\alpha$ denotes alpha matte, $\alpha \cdot F$ is the extracted motion-blurred hand.}
\end{figure}

An image $I$, which contains a motion-blurred hand, is made up by combining foreground hand $F$ and background $B$: $I=\alpha \cdot F+(1-\alpha)\cdot B,\alpha \in[0,1]$, where $\alpha$ is called alpha matte. As shown in this equation, both $\alpha$ and $F$ are related to motion-blurred hands: $\alpha$ demonstrates their transparency; $F$ demonstrates their color. The proposed method separates the task of extracting semi-transparent motion-blurred hand into two subtasks: alpha matte prediction and foreground prediction. The extracted hands can be calculated by multiplying the predicted $\alpha$ and $F$. A simple flow chart of the proposed method is shown in Fig. 1 (b).

Alpha mattes can be calculated by matting methods [7-11]. However, most of matting methods need additional information, such as user supplied trimaps and scribbles [7-11]. In order to avoid the need for user interactions, several matting methods [12-14] employ CNN networks to predict trimaps explicitly or implicitly. The above matting methods focus on static objects, rather than motion-blurred objects. Zhao et. al. [6] proposed a motion-blurred object matting network which only used RGB images as inputs and directly outputted the predicted alpha mattes. In this paper, our alpha matte prediction network is developed from the matting network in [6] by adding a perceptual loss [15].

\begin{figure*}[h]
\centering
\centerline{\includegraphics[width=16 cm]{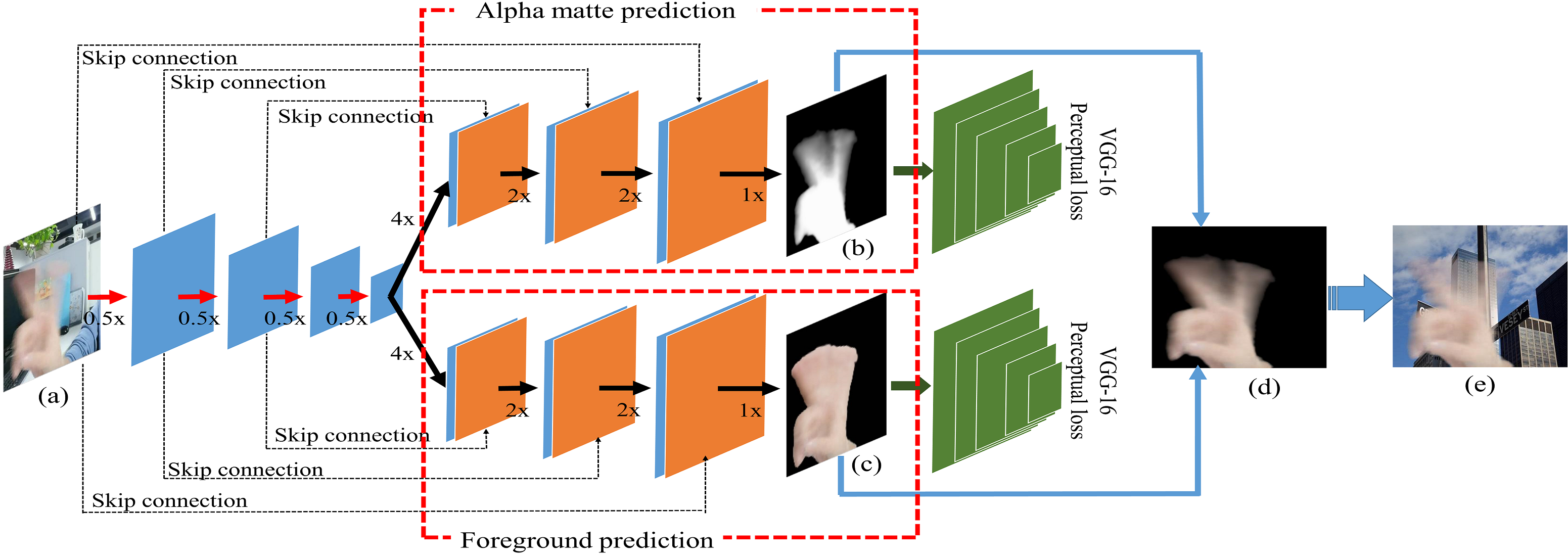}}
\caption{  The Network architecture of the proposed method, where the alpha matte prediction network and foreground prediction network share the same Xception [16] based encoder and have different decoders. In this figure,  (a) is the original image; (b) is the predicted alpha matte; (c) is the predicted foreground; (d) is the extracted motion-blurred hand which is calculated by multiplying (b) and (c); (e) is new image with new background. In this figure, $0.5\times$,  $4\times$, $2\times$, $1\times$  denote the upsampling and downsampling ratios.}
\end{figure*}

Up to now, most of matting methods concentrate on alpha matte prediction. Very few methods can predict foreground images. A recently proposed sampling- and learning-based matting method [11] can estimate the foreground color of unknown regions. However, this method needs user supplied trimaps to annotate the background, foreground, and unknown regions respectively. Besides, this method focuses on static objects, rather than motion-blurred object. In this paper, a network which can automatically predict the foreground images of motion-blurred hands is proposed. The foreground prediction network is a Xception [16] based encoder-decoder network. It only uses the original RGB images as inputs and directly outputs the predicted foreground images. During training, L1-loss and perceptual loss [15] are employed. As shown in Fig. 2, the proposed foreground prediction network shares the same encoder with alpha matte prediction network. But these two prediction networks have different decoders.

Human matting methods [12, 13] extracted human just by $\alpha \cdot I$, rather than $\alpha \cdot F$. A human image $I$ can also be formulated as $I=\alpha \cdot F+(1-\alpha)\cdot B$. Therefore, in semi-transparent areas where $0<\alpha<1$, $\alpha \cdot I$ still contains background information. However, in human images, almost all of the pixels, whose alpha values are between 0 and 1, are located at isolated hairs. Isolated hairs are very thin and have low transparency. Thus, in human images without motion blur, the background information contained in $\alpha \cdot I$ is not obvious and $\alpha \cdot I \approx \alpha\cdot F$. In contrast, in motion-blurred hand images, the areas with high transparency are large. If we extract motion-blurred hands by $\alpha \cdot I$, the extracted hand images will contain obvious background information, which can greatly reduce the sense of reality when changing background. Thus,  motion-blurred hands should be calculated by $\alpha \cdot F$, rather than $\alpha \cdot I$.

In summary, the main contributions in this paper are:

(1)A novel framework which can automatically extract the semi-transparent motion-blurred object from a single image is proposed. This framework consists of two main parts: alpha matte prediction and foreground prediction. The images of extracted objects are calculated by multiplying the predicted alpha mattes and foreground images.

(2)The proposed framework is employed in semi-transparent motion-blurred hand extraction task. Alpha matte prediction and foreground prediction are implemented by Xception-based encoder-decoder networks and trained by L1-loss and perceptual loss.

(3)We enlarge the synthetic motion-blurred hand dataset proposed in [6], and use the enlarged dataset to train the proposed model. Then we use the trained model to process real videos. Experiments demonstrate that the proposed method has promising performance.

\section{Method}

The architecture of the proposed method is shown in Fig. 2. As shown in this figure, the CNN network outputs two kinds of results: predicted alpha mattes and predicted foreground images. The images of extracted hands are calculated by multiplying these two kinds of results.

\subsection{Network architecture}

Encoder-decoder networks have demonstrated their great performance on many pixel-to-pixel prediction tasks, such as segmentation[17-19], depth prediction[20], matting[9, 21], and so on. Encoder-decoder networks usually employ pre-trained image recognition networks, such as VGG[22], ResNet[23], and Xception[16], as the backbone of encoder, and employ several upsampling blocks as decoder. Previous article [17] has shown that encoder-decoder networks based on Xception have better performance and faster speed. Therefore, in this paper, we employ Xception based encoder-decoder networks for alpha matte prediction and foreground prediction. As shown in Fig. 2, the proposed alpha matte prediction network and foreground prediction network share the same encoder, which contains 4 downsampling steps. In each step, the downsampling ratio equals to 0.5. These two prediction networks have two independent decoders. Each decoder contains 3 upsampling steps. The upsampling ratios equal to 4, 2, 2 respectively. In each upsampling step, skip connection is used to recover spatial information. The decoders for alpha matte prediction and foreground prediction have similar structure. But the output of alpha prediction decoder has one channel, while the output of foreground prediction decoder has three channels, which are red channel, green channel, and blue channel respectively.

\subsection{Loss function}

For most of pixel-to-pixel prediction methods, including matting methods [9-11, 13, 14], pixel-wise losses, such as pixel-wise L1-loss and L2-loss, are generally used. However, pixel-wise loss ignores the correlation among pixels. A solution to this problem is employing Conditional Random Fields (CRF) [24-26]. However CRF runs slowly. Another solution is employing perceptual loss. Perceptual loss has been successfully used for style transfer and super-resolution [15, 27]. It calculates the differences between high-level features extracted from predicted images and groundtruth images, and minimizes these differences by backpropagation. High-level features can be extracted by pre-trained convolutional networks.

In this paper, the overall loss $L^o=L^\alpha+L^f$, where $L^\alpha$ denotes alpha prediction loss, $L^f$ denotes foreground prediction loss. Both $L^\alpha$ and $L^f$ are made up by combining L1 losses and perceptual losses.

\subsubsection{Alpha prediction loss}

The alpha prediction loss used in this paper contains three parts: alpha absolute loss $l_{ab}^\alpha$, alpha compositional loss $l_c^\alpha$, and alpha perceptual loss $l_p^\alpha$. $l_{ab}^\alpha$ is the L1 loss between predicted alpha mattes and groundtruth alpha mattes. $l_c^\alpha$ is the L1 loss between predicted compositional images and groundtruth compositional images. Compositional images are generated by $I_c=\alpha_\star \cdot F+(1-\alpha_\star)\cdot B$, where $F$ and $B$ are given foreground images and background images, $\alpha_\star$ denotes predicted alpha mattes or groundtruth alpha mattes.

The alpha perceptual loss $l_p^\alpha$ calculates the L2 loss between high-level features extracted from predicted alpha mattes and groundtruth alpha mattes. In our experiments, VGG-16 [22], which is pretrained for image recognition and contains 5 convolutional blocks, is used as the feature extractor. All of the 5 level feature maps extracted by these 5 convolutional blocks are used to calculate perceptual loss. It should be mentioned that VGG-16 is just employed as a feature extractor, which is only used during training.

The overall alpha prediction loss $L^\alpha=\lambda_{ab}^\alpha l_{ab}^\alpha+\lambda_c^\alpha l_c^\alpha+\lambda_p^\alpha l_p^\alpha$, where $\lambda_{ab}^\alpha$, $\lambda_{c}^\alpha$, and $\lambda_p^\alpha$ are the loss weights. In our experiments, $\lambda_{ab}^\alpha$ and $\lambda_c^\alpha$ are set to 0.5. $\lambda_p^\alpha$ is set to 0.001.

\subsubsection{Foreground prediction loss}

The foreground prediction loss contains two parts: foreground absolute loss $l_{ab}^f$  and foreground perceptual loss $l_p^f$. $l_{ab}^f$ is the  L1 loss between predicted foreground images and groundtruth foreground images. $l_p^f$ is the L2 loss between high-level features extracted from predicted foreground images and groundtruth foreground images. For $l_p^f$, VGG-16 is also used as the feature extractor and all of its 5 levels of feature maps are employed. The overall foreground prediction loss $L^f=\lambda_{ab}^f l_{ab}^f+\lambda_p^f l_p^f$, where $\lambda_{ab}^f$ and $\lambda_p^f$ are the loss weights. In our experiment, they are set to 1 and 0.001 respectively.

\section{Experiment}

In motion-blurred object extraction task, it is almost impossible for human to assign accurate alpha value and foreground color to each image pixel. Therefore, it is very difficult to generate real dataset to train the proposed motion-blurred hand extraction model. In order to solve this problem, synthetic dataset provided by [6] is employed. Besides, in order to increase the diversity of skin colors,  we enlarge this dataset by the same synthetic dataset generation method in  [6], which can generate synthetic motion-blurred hand images, and their corresponding alpha mattes and foreground images. In our experiment, we generate 3 synthetic datasets: training dataset, validation dataset, and testing dataset. These 3 datasets contain 30279, 8283, and 10140 cases respectively. In this paper, all of our models are trained on the synthetic training dataset. The synthetic validation dataset is used to monitor the training processes. The synthetic testing dataset is used to evaluate the performance of our trained models. Then, according to the evaluation results on synthetic testing dataset, the best performing model is chosen to process real videos. In this paper, real videos are captured by a camera on a mobile phone.

Our experiments are implemented under tensorflow with one Nvidia RTX 2080ti GPU and one Intel Core i7 9700k CPU.

\subsection{Evaluation on synthetic testing dataset}

\begin{table}
\caption{Evaluation scores on synthetic testing dataset. SAD is short for sum of absolute differences. MSE is short for mean squared error. PL is short for perceptual loss. SE is short for sharing encoder.}
\label{table}
\small
\setlength{\tabcolsep}{3pt}
\centering
\begin{tabular}{|p{50pt}|p{30pt}|p{30pt}|p{50pt}|p{50pt}|}
\hline
\multicolumn{5}{|c|}{Alpha prediction} \\
\hline
 Methods & SE & PL & SAD$(\times10^3)$ &  MSE$(\times10^{-3})$ \\
\hline
Model 1 & no  & no  & 2.57 & 1.29  \\
Model 2 & yes & no  & 2.47 & 1.20  \\
Model 3 & no  & yes & 2.08 & 0.73  \\
Model 4 & yes & yes & {\bf 2.02}  & {\bf 0.71} \\
\hline
\multicolumn{5}{|c|}{Foreground prediction} \\
\hline
 Methods & SE & PL & SAD$(\times10^3)$ &  MSE$(\times10^{-3})$ \\
\hline
Model 1 & no  & no  & 8.84 & 2.30  \\
Model 2 & yes & no  & 8.73 & 2.06  \\
Model 3 & no  & yes & 8.08 & 1.71  \\
Model 4 & yes & yes & {\bf 7.94}  & {\bf 1.58} \\
\hline
\end{tabular}
\label{tab1}
\end{table}

\begin{figure}
\centerline{\includegraphics[width=\columnwidth]{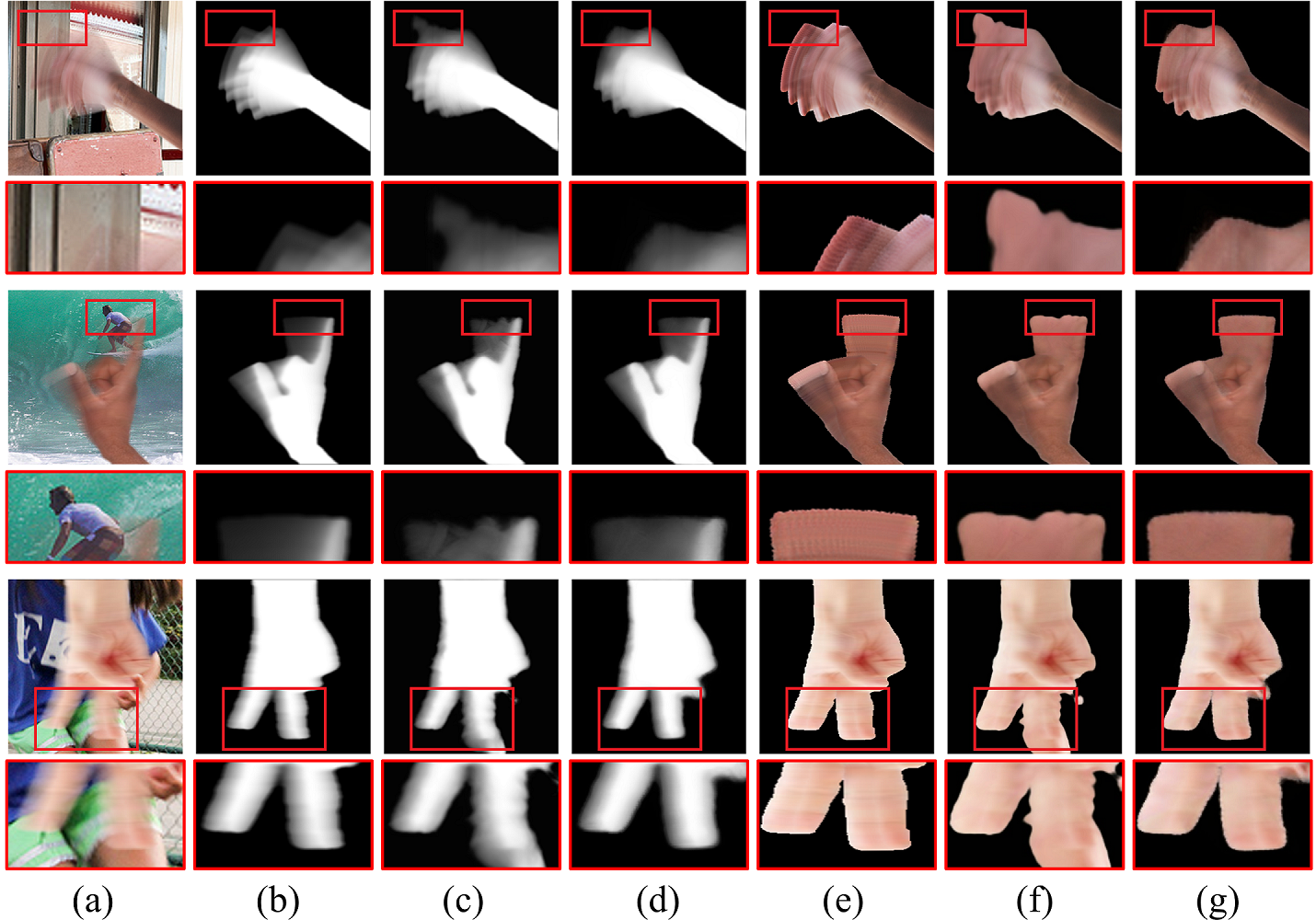}}
\caption{Compare the results predicted by the networks trained without and with perceptual loss. (a) original images. (b) groundturth alpha mattes. (c) alpha mattes predicted by Model 2(without perceptual loss). (d) alpha mattes predicted by Model 4(with perceptual loss). (e) groundtruth foreground images. (f) foreground images predicted by Model 2(without perceptual loss). (g) foreground images predicted by Model 4(with perceptual loss).}
\end{figure}

In this paper, four models are trained. These four models are called as Model 1, Model 2, Model 3, and Model 4 respectively. Particularly, Model 1 is trained without perceptual loss and its two subtasks do not share the same encoder; Model 2 is trained without perceptual loss and its two subtasks share the same encoder; Model 3 is trained with perceptual loss and its two subtasks do not share the same encoder; Model 4 is trained with perceptual loss and its two subtasks share the same encoder. The alpha matte prediction network of Model 1 is the same matting network proposed in [6]. The evaluation scores on the synthetic testing dataset are shown in Table 1. As shown in this table, Model 4 has the best performance. Therefore, Model 4 is chosen to process real videos in next subsection. In our experiment, during training Model 4, the learning rate was set to $3.5\times10^{-3}$; the momentum was set to 0.9; the weight decay was set to $4\times10^{-5}$. The training process ended after 80 epoches.

As shown in Table 1, sharing the same encoder can slightly improve the prediction performance, while adding perceptual losses can obviously improve the prediction performance. In order to show the effectiveness of perceptual loss qualitatively, several alpha prediction and foreground prediction examples of Model 2 and Model 4 are shown in Fig. 3. Model 2 and Model 4 have the same network structure  and are trained without and with perceptual loss respectively. Fig. 3 shows that the results predicted by Model 4 have more accurate and reasonable shapes and textures than Model 2.

\subsection{Experiments on real videos}

As described in last subsection, the best-performing model, Model 4, is chosen to predict the alpha mattes and foreground images of motion-blurred hands in real videos. The extracted hand images are calculated by multiplying the predicted alpha mattes and foreground images. Because it is very hard to generate groundtruth for real datasets, in this subsection, we just show and compare the motion-blurred hand extraction performances qualitatively.

\subsubsection{Compare with other methods}

\begin{figure}
\centerline{\includegraphics[width=\columnwidth]{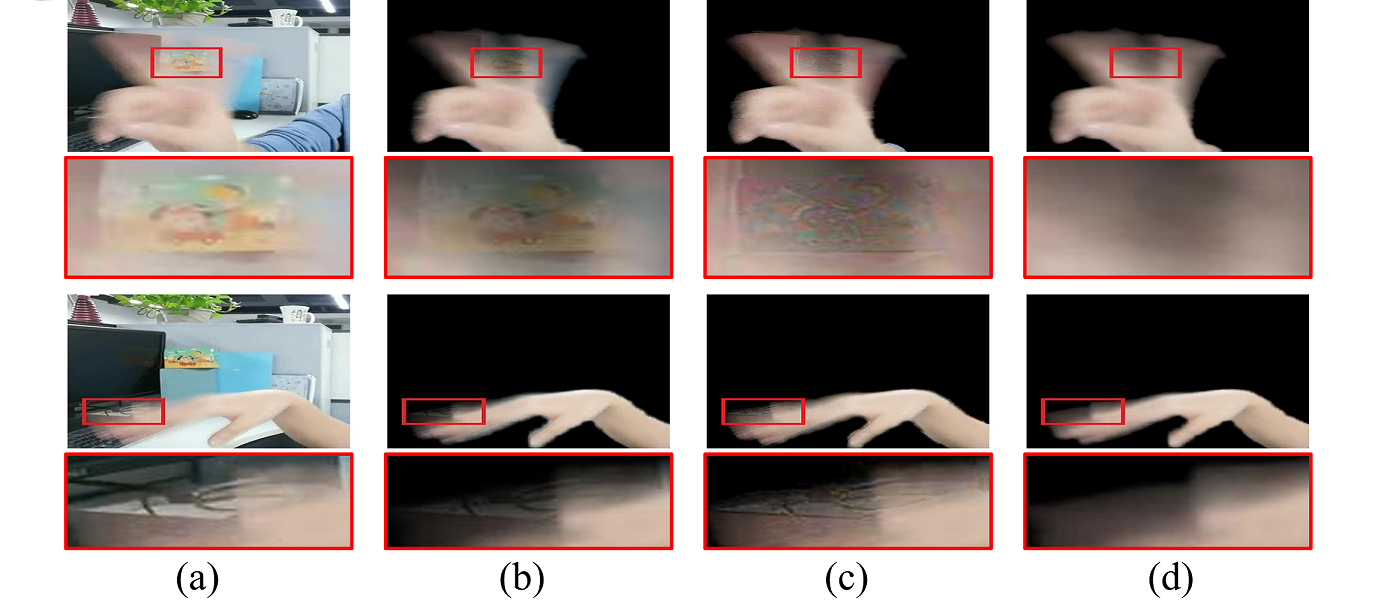}}
\caption{Compare the hand extraction performance of different methods. (a) original RGB frames. (b) hand images extracted by multiplying the predicted alpha mattes and original frames. This strategy is popular used in human extraction methods [12,13]. (c) hand images extracted by subtracting the background component from the original frames. This method is employed in [6]. (d) hand images extracted by the method proposed in this paper.}
\end{figure}

The proposed motion-blurred hand extraction method is compared with other two previous methods. One method extracts motion-blurred hands simply by multiplying the predicted alpha mattes and original frames. This strategy is popularly used in human extraction methods [12, 13]. The other method extracts motion-blurred hands by subtracting the background component from the original frames [6]. Its flow chart is shown Fig.1 (a). In the following, the above two methods are called as Method 1 and Method 2 for convenience. As shown in Fig. 4, the hand images extracted by Method 1 contains obvious background information, and the hand images extracted by Method 2 contains obvious color distortions. In contract, the hand images generated by our method are smooth and natural. In addition, Method 2 needs background images, which are not convenient to obtain, while our method doesn't need any additional information. Therefore, our method has the best performance and can be applied more widely.

\subsubsection{Modify human segmentation results}

\begin{figure}
\centerline{\includegraphics[width=\columnwidth]{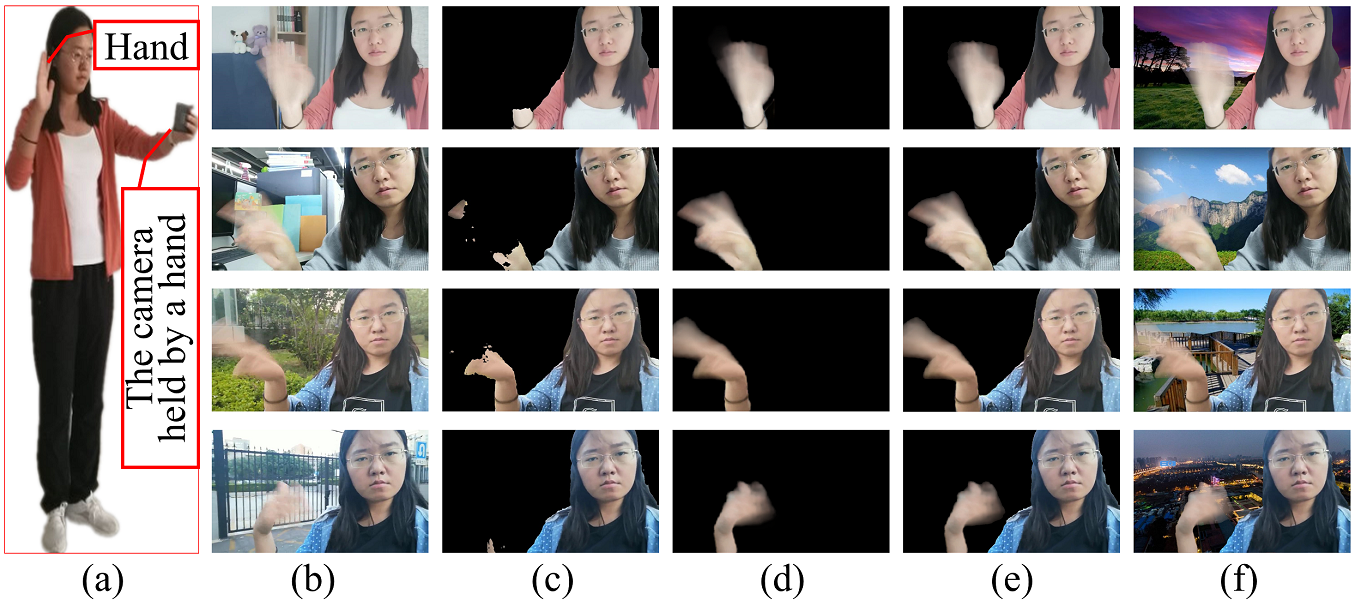}}
\caption{Some examples of modifying human segmentation results. (a) a picture shows how we capture our videos. (b) original RGB frames. (c) human segmentation results predicted by state-of-art image segmentation method Deeplab v3+ [17]. (d) the motion-blurred hand images extracted by the proposed method. (e) the modified human soft segmentation results. (f) new frames with new background.  }
\end{figure}

In previous work [6], the extracted motion-blurred hand images were used to modify human segmentation results. In this paper, we also do the same job, in order to verify that the proposed motion-blurred hand extraction method is useful in practical applications. The method used in [6] needs background images. Therefore, in [6], videos were captured by static cameras in order to obtain background images from the nearby frames. In contrast, the method proposed in this paper doesn’t need background images and can be used in videos captured by moving cameras. The examples shown in Fig. 5 are captured by a hand-held camera. Particularly, we hold camera by one hand and make hand gestures by the other hand. In Fig. 5, the 4 examples are captured at home, at workplace, in a park, and near a road respectively. These examples demonstrate that state-of-art segmentation method can’t deal with motion-blurred hands, and our motion-blurred hand extraction method has good performance to modify human segmentation results and change background.

\section{CONCLUSION AND FUTURE WORK}

In this paper, a novel method which can automatically extract the semi-transparent motion-blurred hand from a single image is proposed. This novel method contains two main parts: alpha matte prediction and foreground prediction. The images of extracted hands are calculated by multiplying the predicted alpha mattes and foreground images. In the future, we will try to find an efficient method which can extract complete motion-blurred objects, such as motion-blurred human.

\end{document}